%% file: main.tex
\documentclass[12pt]{article}
\usepackage[utf8]{inputenc}
\usepackage{amsmath}
\usepackage{amssymb}
\usepackage{natbib}

\usepackage{fancyhdr}
\usepackage{longtable}
\usepackage{ulem}

\usepackage[table,xcdraw, dvipsnames]{xcolor}

\usepackage[font=small,labelfont=bf,labelsep=space]{caption}

\usepackage{xspace}
\usepackage[pdftex]{graphicx}
\usepackage{hyperref}
\usepackage[capitalize]{cleveref}

\hypersetup{
    colorlinks = true,
    linkcolor=ACMDarkBlue,
    citecolor=ACMDarkBlue,
    urlcolor=ACMDarkBlue
}
\definecolor[named]{ACMDarkBlue}{cmyk}{1,0.58,0,0.21}

\usepackage{authblk}
\usepackage{geometry}
\usepackage{parskip} 
\usepackage[inline]{enumitem}

\usepackage{booktabs}  
\usepackage{amsfonts}
\usepackage{mathtools}
\usepackage{url}
\usepackage{enumitem}
\usepackage{comment}
\usepackage{mdframed}
\usepackage{wrapfig}
\usepackage{floatrow}
\usepackage[nottoc]{tocbibind}
\renewcommand\bibname{References}

\usepackage{ntheorem}
\newtheorem{theorem}{Theorem}[section]

\newtheorem{hypo}[theorem]{Hypothesis}
\newtheorem{prop}[theorem]{Proposition}

\definecolor{Gray1}{gray}{0.82}
\definecolor{Gray2}{gray}{0.92}

\pdfmapfile{=Cochineal.map}
\usepackage{cochineal}
\usepackage[T1]{fontenc}

\usepackage[scale=.95,type1]{cabin}
\usepackage[zerostyle=c,scaled=.94]{newtxtt}
\usepackage[cal=boondoxo]{mathalfa}
\usepackage{microtype}

\usepackage{etoolbox}

\input{macros}

\usepackage{etoolbox}

\usepackage[dvipsnames]{xcolor}  
\definecolor{Blue4Head}{HTML}{004488} 
\fancypagestyle{plain}{%
  \fancyhf{}
  \fancyhead[L]{
    \raisebox{-0.15\height}{\includegraphics[height=1.2\baselineskip]{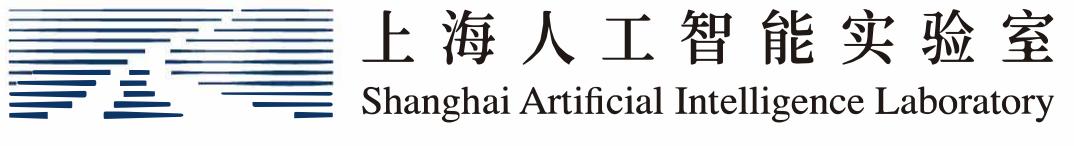}}%
  }
  \fancyhead[R]{
    \raisebox{-0.15\height}{\includegraphics[height=1\baselineskip]{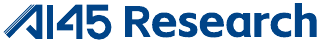}}%
  }
  
  \fancyfoot[C]{\thepage}
}

\fancyheadoffset{0pt} 
\fancyfootoffset{0pt} 

\usepackage{chngcntr} 
\counterwithout{figure}{section}

\title{\bf \texttt{R$^\textbf{2}$AI}: Towards Resistant and Resilient AI\\ in an Evolving World}

\author[1,2]{Youbang Sun}
\author[1,3]{Xiang Wang}
\author[1]{Jie Fu}
\author[1]{Chaochao Lu}
\author[1,2]{Bowen Zhou}

\affil[1]{Shanghai Artificial Intelligence Laboratory}
\affil[2]{Tsinghua University}
\affil[3]{University of Science and Technology of China}

{\small\date{September 4, 2025}}
\begin{document}

\maketitle
  
\pagestyle{fancy}
\renewcommand{\thefootnote}{}

\renewcommand{\thefootnote}{\arabic{footnote}}
\setcounter{footnote}{0}

\input{chapters/0_abstract}
\input{chapters/1_introduction}

\input{chapters/2_rethinking}

\input{chapters/3_paradigm}
\input{chapters/4_path}

\input{chapters/5_implications}

\input{chapters/6_conclusion_ack}

\bibliographystyle{bibsty}	
\bibliography{reference}

\end{document}

%% file: macros.tex
\apptocmd{\thebibliography}{\raggedright}{}{}

\usepackage{multirow}
\usepackage{tablefootnote}
\usepackage{algpseudocode}
\usepackage{algorithm}
\usepackage[pdftex]{graphicx}
\usepackage{makecell}
\usepackage{tabularx}
\usepackage{subfigure}  
\usepackage{graphicx}
\usepackage{enumitem}
\usepackage{CJKutf8}
\CJKtilde

\usepackage{bera}
\usepackage{listings}
\usepackage{xcolor}
\definecolor{eclipseStrings}{RGB}{42,0.0,255}
\lstdefinelanguage{json}{
    basicstyle=\scriptsize\ttfamily,
    commentstyle=\color{eclipseStrings},
    showstringspaces=false,
    breaklines=true,
    frame=single,
    rulecolor=\color{black},
    string=[s]{"}{"},
    comment=[l]{:\ "},
    morecomment=[l]{:"},
    literate=
        *{0}{{{\color{numb}0}}}{1}
         {1}{{{\color{numb}1}}}{1}
         {2}{{{\color{numb}2}}}{1}
         {3}{{{\color{numb}3}}}{1}
         {4}{{{\color{numb}4}}}{1}
         {5}{{{\color{numb}5}}}{1}
         {6}{{{\color{numb}6}}}{1}
         {7}{{{\color{numb}7}}}{1}
         {8}{{{\color{numb}8}}}{1}
         {9}{{{\color{numb}9}}}{1}
}

\numberwithin{figure}{section}
\numberwithin{table}{section}

\usepackage{pgfplotstable} 
\usepackage{tikz}    
\usepackage{placeins}

\theoremseparator{:} 

\newmdtheoremenv[%
  backgroundcolor=white,
  linecolor=blue!60!black,
  linewidth=2pt,
  topline=true,
  rightline=false,
  skipabove=10pt,
  skipbelow=10pt,
  leftline=false]{ourexample}{Application}
  
\newmdtheoremenv[%
  backgroundcolor=gray!20,
  linecolor=red!60!black,
  linewidth=2pt,
  topline=false,
  rightline=false,
  skipabove=10pt,
  skipbelow=10pt,
  leftline=false]{ourbox}{Formulation}

\newmdtheoremenv[%
  backgroundcolor=gray!20,
  linecolor=red!60!black,
  linewidth=2pt,
  topline=false,
  rightline=false,
  skipabove=10pt,
  skipbelow=10pt,
  leftline=false]{regbox}{Box}

\theoremstyle{nonumberplain}
\newmdtheoremenv[%
  backgroundcolor=gray!20,
  linecolor=red!60!black,
  linewidth=2pt,
  topline=false,
  rightline=false,
  skipabove=10pt,
  skipbelow=10pt,
  leftline=false]{suppregbox}{Box S1}

\newcommand{\eg}{\textit{e.g.,}\xspace}

\definecolor{quotemark}{gray}{0.7}
\makeatletter
\def\fquote{%
    \@ifnextchar[{\fquote@i}{\fquote@i[]}
           }%
\def\fquote@i[#1]{%
    \def\tempa{#1}%
    \@ifnextchar[{\fquote@ii}{\fquote@ii[]}
                 }%
\def\fquote@ii[#1]{%
    \def\tempb{#1}%
    \@ifnextchar[{\fquote@iii}{\fquote@iii[]}
                      }%
\def\fquote@iii[#1]{%
    \def\tempc{#1}%
    \vspace{1em}%
    \noindent%
    \begin{list}{}{%
         \setlength{\leftmargin}{0.1\textwidth}%
         \setlength{\rightmargin}{0.1\textwidth}%
                  }%
         \item[]%
         \begin{picture}(0,0)%
         \put(-15,-5){\makebox(0,0){\scalebox{3}{\textcolor{quotemark}{``}}}}%
         \end{picture}%
         \begingroup\itshape}%
 \def\endfquote{%
 \endgroup\par%
 \makebox[0pt][l]{%
 \hspace{0.8\textwidth}%
 \begin{picture}(0,0)(0,0)%
 \put(15,15){\makebox(0,0){%
 \scalebox{3}{\color{quotemark}''}}}%
 \end{picture}}%
 \ifx\tempa\empty%
 \else%
    \ifx\tempc\empty%
       \hfill\rule{100pt}{0.5pt}\\\mbox{}\hfill\tempa,\ \emph{\tempb}%
   \else%
       \hfill\rule{100pt}{0.5pt}\\\mbox{}\hfill\tempa,\ \emph{\tempb},\ \tempc%
   \fi\fi\par%
   \vspace{0.5em}%
 \end{list}%
 }%
 \makeatother

%% file: chapters/0_abstract.tex
\begin{abstract}
\noindent
In this position paper, we address the persistent gap between rapidly growing AI capabilities and lagging safety progress. Existing paradigms divide into ``Make AI Safe'', which applies post-hoc alignment and guardrails but remains brittle and reactive, and ``Make Safe AI'', which emphasizes intrinsic safety but struggles to address unforeseen risks in open-ended environments. We therefore propose \textit{safe-by-coevolution} as a new formulation of the ``Make Safe AI'' paradigm, inspired by biological immunity, in which safety becomes a dynamic, adversarial, and ongoing learning process. To operationalize this vision, we introduce \texttt{R$^2$AI}---\textit{Resistant and Resilient AI}---as a practical framework that unites resistance against known threats with resilience to unforeseen risks. \texttt{R$^2$AI} integrates \textit{fast and slow safe models}, adversarial simulation and verification through a \textit{safety wind tunnel}, and continual feedback loops that guide safety and capability to coevolve. We argue that this framework offers a scalable and proactive path to maintain continual safety in dynamic environments, addressing both near-term vulnerabilities and long-term existential risks as AI advances toward AGI and ASI.
\end{abstract}

%% file: chapters/1_introduction.tex
\section{Introduction}

Recent years have witnessed rapid developments and huge breakthroughs in AI, leading to its integration into everyday life and establishing it as a foundational infrastructure in society \citep{van2024big}.
As AI systems are increasingly deployed in safety-critical domains (\eg scientific research \citep{alphafold,zhang2023artificial,alphaevolve}, autonomous driving \citep{AdvSim,CtRL-Sim}, healthcare \citep{panayides2020ai, bekbolatova2024transformative}, law \citep{lai2024large}), the risks posed by unsafe or unreliable outputs have become more pronounced.
In such settings, failures can result in severe, even catastrophic, consequences.
Beyond these near-term concerns, the continued advancement toward highly autonomous and superhuman-level AI raises long-term existential risks \citep{dalrymple2024towards,bengio2025superintelligent,bengio2025singapore,kulveit2025position,clymer2025bare,lab2025frontier}.
As capabilities scale, so does the difficulty of aligning, controlling, and governing these systems, thus potentially leading to scenarios with irreversible societal or civilizational impacts \citep{bengio2025international,shlab2025safework_f1_framework}.

\begin{figure}[t]
    \centering
    \includegraphics[width=\linewidth]{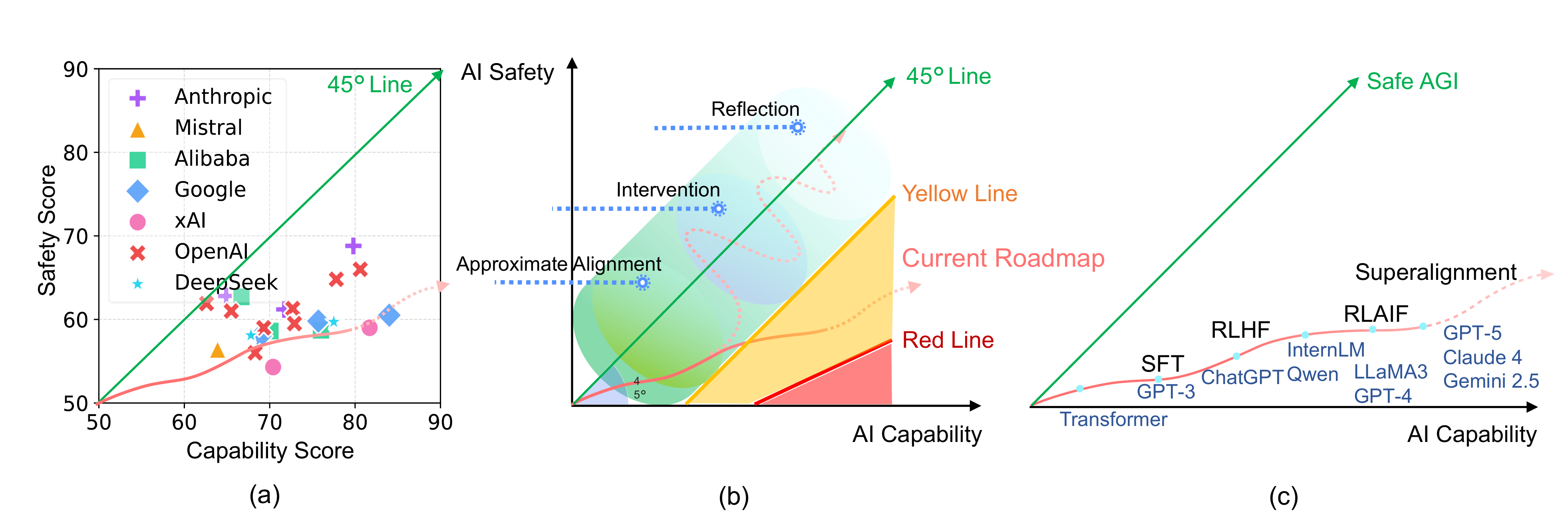}
    \caption{The AI-45$^{\circ}$ Law \citep{yang2024towards}: coevolving capability with safety.\protect\footnotemark (a) Empirical distribution of leading foundation models, showing a widening gap between capability scores and safety scores across major labs. (b) Conceptual safety–capability plane comparing the current roadmap (pink) with the yellow, red, and 45$^{\circ}$ trajectories toward safe AGI, emphasizing transitions from approximate alignment to reflection. (c) Historical timeline of frontier models, from Transformer \citep{vaswani2017attention} to GPT-5 \citep{openai-gpt5-2025}, Claude-4 \citep{anthropic_claude4}, and Gemini-2.5 \citep{comanici2025gemini}, illustrating the divergence between capability scaling and current alignment methods (\eg SFT \citep{ouyang2022training}, RLHF \citep{christiano2017deep}, RLAIF \citep{bai2022constitutional}), and the need for a coevolutionary path to Safe AGI.}
    \label{fig:AI45}
\end{figure}
\footnotetext{Figure \ref{fig:AI45}\textcolor{ACMDarkBlue}{a} is reproduced from data available at \url{https://aiben.ch}. Figures \ref{fig:AI45}\textcolor{ACMDarkBlue}{b} and \ref{fig:AI45}\textcolor{ACMDarkBlue}{c} are adapted from Figure 1 in \citet{yang2024towards}.}

Despite escalating risks, safety progress has lagged far behind capability growth. As shown in Figure \ref{fig:AI45}\textcolor{ACMDarkBlue}{a}, evaluations show a consistent pattern: leading AI models worldwide---such as GPT-5 \citep{openai-gpt5-2025}, Claude 4 \citep{anthropic_claude4}, and Gemini-2.5 \citep{comanici2025gemini}---demonstrate significantly higher capability scores than safety scores. This imbalance reveals a structural problem: current safety approaches are reactive, fragmented, and incapable of scaling with capability. To capture this tension, Shanghai AI Lab proposed the AI-45$^{\circ}$ Law \citep{yang2024towards}: safety and capability must coevolve along a 45$^{\circ}$ diagonal trajectory. Temporary deviations are tolerable, but persistent dips below the 45° line increase the risk of catastrophic misalignment, while rising above it may unnecessarily stall innovation. We further define two thresholds: \textit{yellow lines} serve as early warnings when capability begins to outpace safety; \textit{red lines} denote irreversible, catastrophic risks that must never be crossed \citep{idais2024statement,idais2025statement}.

Current safety research can be broadly categorized into two paradigms. The dominant ``Make AI Safe'' paradigm seeks to improve safety after model development, typically through alignment fine-tuning (\eg RLHF \citep{christiano2017deep}, RLAIF \citep{bai2022constitutional}), red teaming \citep{perez2022red,ganguli2022red,pavlova2024automated}, and guardrail \citep{bai2022constitutional,shreya2023guardrails,oh2024uniguard}. While effective in mitigating known risks, these methods are often reactive, brittle, expensive, and struggle to address unknown or emerging risks. In contrast, the ``Make Safe AI'' paradigm emphasizes intrinsic safety, designing systems to be safe by construction. Prominent directions include formal guarantees \citep{szegedy2020promising, dalrymple2024towards} and Scientist AI \citep{bengio2025superintelligent}. Yet even these approaches often fall short in open-ended environments where novel risks cannot be fully anticipated.

To achieve \textit{scalable safety} in an envolving world, we must rethink what ``Make Safe AI'' entails. We argue that its core principle should be \textit{coevolution}: safety must not be treated as a constraint or one-time guarantee, but as a continuous, adaptive capability that evolves alongside intelligence in uncertain, dynamic environments. We therefore propose \textit{safe-by-coevolution} as a new formulation for ``Make Safe AI'', inspired by biological immunity \citep{cooper2006evolution,muller2018evolutionary,papkou2019genomic}, in which safety becomes a dynamic, adversarial, and ongoing learning process. By embedding coevolutionary mechanisms into the AI lifecycle, systems can remain safe through sustained interaction with real and simulated environments. Just as human immunity develops through continual exposure to pathogens \citep{flajnik2010origin,nourmohammad2016host,buckingham2022coevolutionary}, AI must develop safety through ongoing interaction with its environment. Without such an ``immune system'', advanced AI risks becoming powerful yet dangerously brittle, and unlike humans, a single catastrophic failure could be irreversible.

Safe-by-coevolution advances a proactive path for safety evolution. It is structured around three iterative steps: 1) \textit{Near-term safety guarantee}: ensure that an AI system at time $t_0$ is verifiably within a defined safety margin; 2) \textit{Safe iterative step}: for any system already safe, design coevolutionary mechanisms---adversarial interactions, feedback loops, and continuous updates---to guide each upgrade back within that margin; 3) \textit{Continual safety by induction}: repeat this loop so that safety evolves in sync with capability. Unlike reactive patching, this approach integrates safety into the developmental process. To address unforeseen risks (\eg paradigm shifts or red-line events), it further incorporates a \textit{reset-and-recover} mechanism: halting unsafe systems, redefining safety margins, and establishing new verified checkpoints to sustain coevolution.

To realize this vision, we introduce \texttt{R$^2$AI}---\textit{Resistant and Resilient AI}---as a practical framework for safe-by-coevolution. \texttt{R$^2$AI} unifies \textit{resistance} and \textit{resilience} as the two foundational and complementary dimensions of intrinsic safety: resistance captures robustness against known threats, while resilience emphasizes recovery and adaptation under unforeseen risks. Specifically, \texttt{R$^2$AI} comprises four interacting components: (i) \textit{fast safe models} for real-time response, (ii) \textit{slow safe models} for verification and reasoning, (iii) a \textit{safety wind tunnel} that simulates adversarial attacks and validation loops, and (iv) an \textit{external environment} for interacting with diverse, realistic scenarios. Through adversarial and cooperative dynamics, these components coevolve to embed safety as a learned and adaptive property. Over time, slow mechanisms become internalized into fast, intuitive safeguards, thereby lowering the cost of compliance and enabling scalable, intrinsic safety even at the frontier of AGI \citep{goertzel2014artificial,lake2017building,baum2017survey,bubeck2023sparks,morris2024position,raman2025navigating}.

\paragraph{Our position.} 
We propose \texttt{R$^2$AI}---a framework uniting resistance and resilience---as a scalable and intrinsically adaptive approach to AI safety. Grounded in \textit{safe-by-coevolution}, it reconceives safety as a continual learning process rather than a static constraint, enabling systems to withstand known threats, adapt to unforeseen risks, and evolve in step with capability. This redefinition offers a generalizable alternative to brittle alignment or top-down control, providing a proactive path to sustain safety across dynamic environments and future ASI \citep{nick2014superintelligence,kim2024road,hendrycks2025superintelligence}.

\paragraph{Structure of this paper.} 

The remainder of the paper is organized as follows. Section~\ref{sec:rethink} reconsiders the paradigm of ``Make Safe AI'' and introduces \textit{resistance} and \textit{resilience} as its foundational properties. Section~\ref{sec:evolve} formalizes the \textit{safe-by-coevolution} principle, establishing its theoretical foundation and operational steps. Section~\ref{sec:R2AI} presents the \texttt{R$^2$AI} framework, detailing its core components, mechanisms, and continual learning strategies. Section~\ref{sec:impact} discusses the implications, applications, and societal impacts of \texttt{R$^2$AI}, highlighting its relevance to both near-term safety challenges and long-term existential risks.

%% file: chapters/2_rethinking.tex
\section{Rethinking ``Make Safe AI''}\label{sec:rethink}

\begin{figure}[t]
    \centering
    \includegraphics[width=\linewidth]{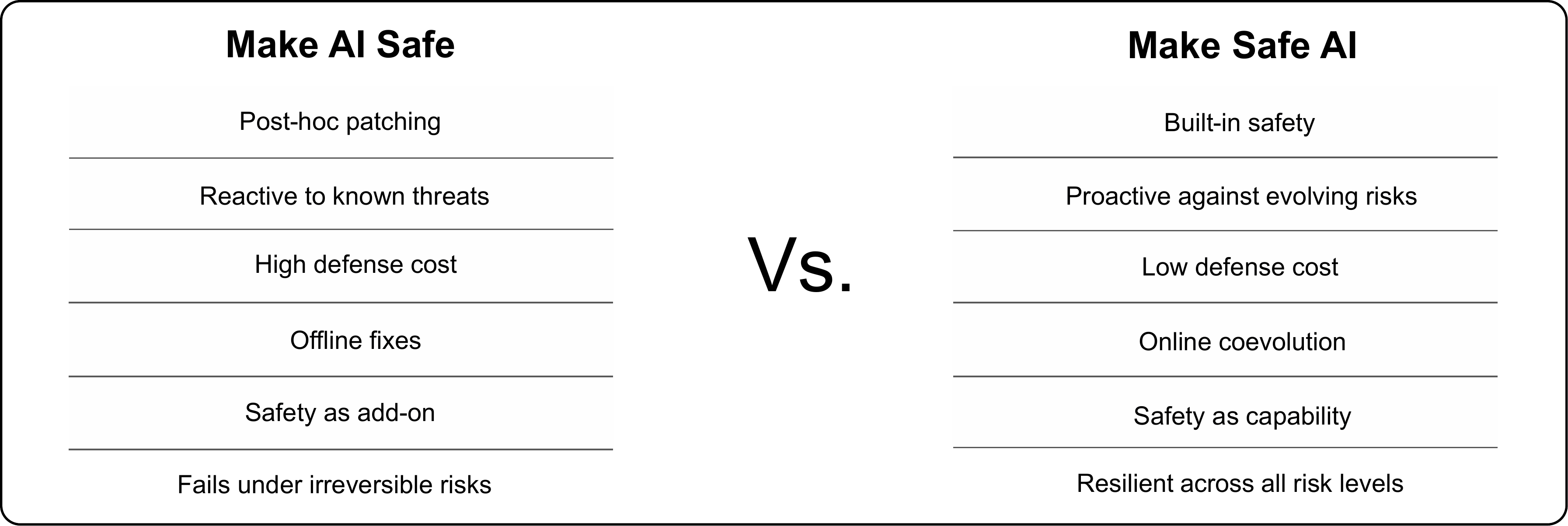}
    \caption{Conceptual contrast between ``Make AI Safe'' and ``Make Safe AI''.}
    \label{fig:safeai}
\end{figure}

The contrast between ``Make AI Safe'' and ``Make Safe AI'', as shown in Figure~\ref{fig:safeai}, underscores a fundamental shift in perspective. While ``Make AI Safe'' relies on post-hoc fixes, reactive defenses, and costly patching that falter under irreversible risks, ``Make Safe AI'' envisions safety as a built-in, proactive, and evolving capability. This transition requires rethinking safety not as an external add-on, but as an intrinsic property that coevolves with intelligence.

Existing work toward ``Make Safe AI'' has made important progress---ranging from formal guarantees \citep{seshia2022toward, dalrymple2024towards} to constrained design choices such as Scientist AI \citep{bengio2025superintelligent} and Tool AI \citep{karnofsky2024if}. Yet these approaches struggle in open-ended, non-stationary environments where novel objectives, adversarial pressures, and distributional shifts are inevitable.

We argue that the foundation of ``Make Safe AI'' must rest on two complementary properties, inspired by ecological systems where long-lived organisms survive under continual stress \citep{holling1973resilience,levin1998ecosystems,gunderson2000ecological,walker2004resilience}: \textit{resistance}, the capacity to withstand and mitigate known threats, and \textit{resilience}, the capacity to recover, adapt, and improve in the face of unforeseen disturbances. Unlike static safeguards, these properties are endogenous, enabling systems to maintain integrity across dynamic and uncertain environments.

Building on this foundation, we propose \textit{safe-by-coevolution} as a new formulation of the ``Make Safe AI'' paradigm. Inspired by biological immunity \citep{cooper2006evolution,muller2018evolutionary,papkou2019genomic}, this principle reconceives safety as a dynamic, adversarial, and ongoing learning process. Rather than attaching fixed safeguards to capable systems, safety itself must scale with capability---reflexively, adaptively, and proactively. This redefinition is essential for ensuring that AI systems preserve both functional integrity and ethical alignment in real-world complexity.

\subsection{Levels of ``Make Safe AI''}

\begin{figure}[t]
    \centering
    \includegraphics[width=\linewidth]{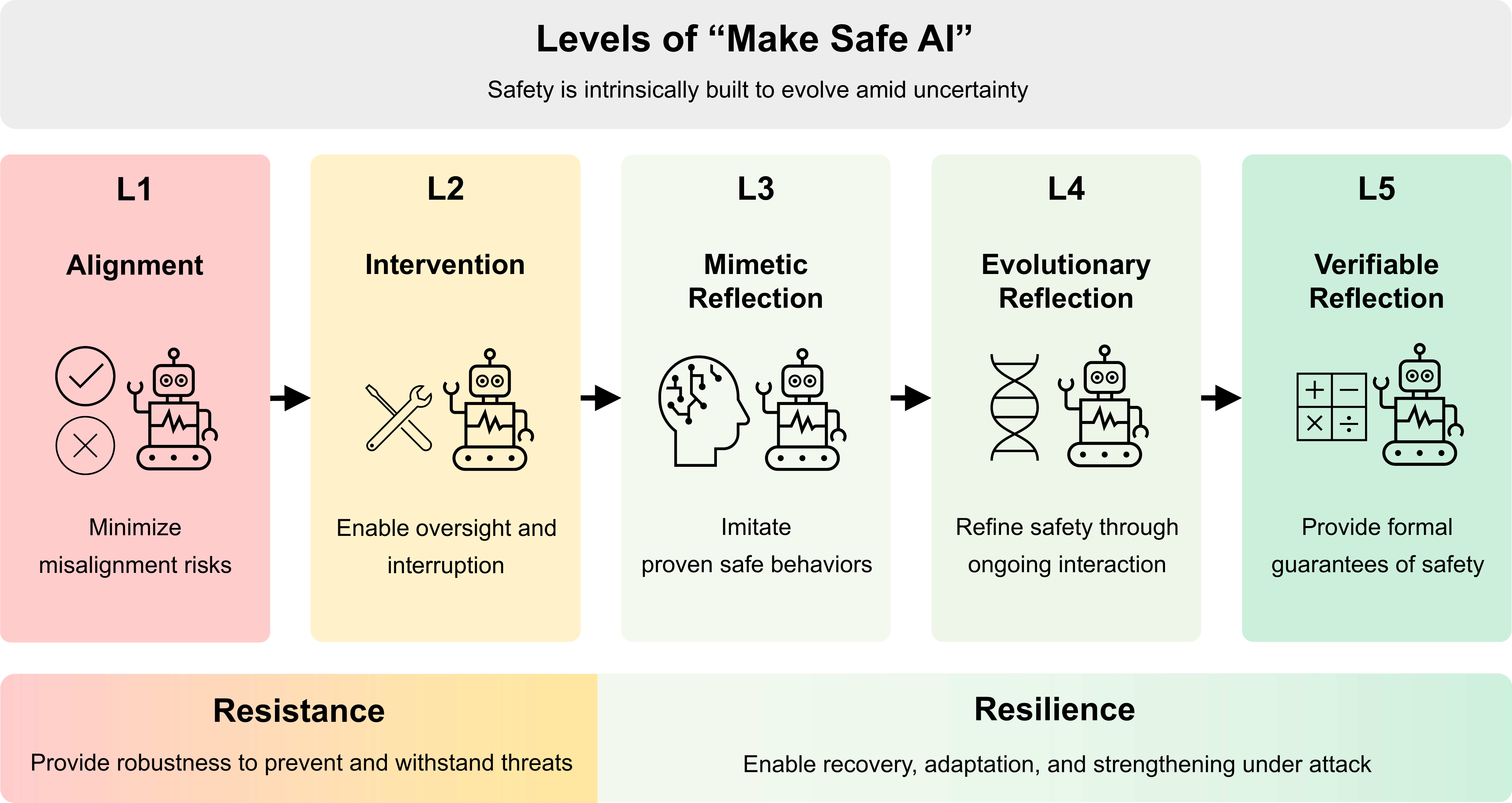}
    \caption{Five levels of ``Make Safe AI'', which progressively embed safety as an intrinsic and evolving capability.}
    \label{fig:levels}
\end{figure}

Building on this redefinition of ``Make Safe AI'', we can further structure its progression into a hierarchy of safety levels. As illustrated in Figure~\ref{fig:levels}, safety is not a binary attribute but an evolving spectrum, intrinsically built to adapt amid uncertainty. This perspective highlights how safety matures from basic alignment toward fully verified, self-evolving guarantees. To instantiate this view, we propose a five-level spectrum that extends the causal ladder of trustworthy AI \citep{yang2024towards}. This spectrum reflects increasing degrees of adaptivity, autonomy, and assurance in dynamic environments, capturing the progression from approximate alignment to formal, verifiable safety.

At the foundational layers, \textit{resistance} anchors safety by providing robustness against known risks: \textit{L1 Alignment} minimizes misalignment through approximate tuning, and \textit{L2 Intervention} ensures oversight and the ability to halt unsafe behavior. Building upward, \textit{resilience} enables adaptive safety beyond reactive correction: \textit{L3 Mimetic Reflection} introduces internal reflection by imitating proven safe behaviors to anticipate risks, \textit{L4 Evolutionary Reflection} advances this reflection into continual co-adaptation with environments, and \textit{L5 Verifiable Reflection} culminates in formalized reflection, where provable guarantees sustain resilience even under uncertainty. Specifically, 

\begin{itemize}
    \item \textbf{L1: Alignment.}
    Safety is achieved through approximate alignment \citep{yang2024towards}, typically via supervised fine-tuning, direct preference optimization \citep{dpo,simpo,betadpo}, reinforcement learning from human feedback \citep{ouyang2022training, bai2022constitutional,grpo}, knowledge editing \citep{memit,alphaedit,anyedit}, or activation steering \citep{arditi2024refusal,panickssery2023steering}. While practical, such alignment is static and correlation-based, providing robustness against known risks but requiring continual updates to withstand new tasks or adversarial strategies \citep{perez2023discovering,zou2023universal,wei2023jailbroken,yi2024jailbreak,ji2024language}.
    
    \item \textbf{L2: Intervention.}
    Safety is treated as a control problem \citep{hendrycks2021unsolved}, where systems monitor outputs and intervene when thresholds are violated \citep{orseau2016safely,Circuit-breakers,SafeDecoding}, guided by explicit feedback \citep{bengio2025international,DBLP:journals/corr/abs-2502-12970}. This level provides oversight and interruption, offering robustness through reactive correction \citep{ganguli2022red}. In addition, advances in mechanistic interpretability \citep{sharkey2025open} provide tools to identify and intervene on unsafe internal circuits or representations before they manifest in outputs \citep{nanda2023progress,conmy2023towards,bereska2024mechanistic}. However, the overall effectiveness depends on timely and reliable feedback signals \citep{leike2018scalable,lin2021truthfulqa,terekhov2025control}.
    
    \item \textbf{L3: Mimetic Reflection.}
    At this level, the system engages in \textit{reflection by imitation}, developing internal reasoning capabilities \citep{shinn2023reflexion,madaan2023self,DBLP:journals/corr/abs-2412-16339,lab2025safework,Backtrackingsafety,DBLP:journals/corr/abs-2501-19180,alphaalign}. It can perform counterfactual reasoning, simulate outcomes, and anticipate risks by internalizing proven safe behaviors \citep{dai2023safe,reddy2024safety}. This marks a shift from externally imposed oversight to internalized safety reasoning, enabling anticipatory resilience and reducing dependence on continuous supervision.
    
    \item \textbf{L4: Evolutionary Reflection.}
    Reflection becomes \textit{evolutionary}: safety mechanisms themselves adapt through continual interaction and coevolution with capabilities and environments \citep{pan2025evo,cai2025aegisllm}. Safety thus becomes an agentic property \citep{lifelongsafety}---self-directed, adaptive, and scalable to complex or unforeseen challenges---enabling recovery and strengthening under attack.
    
    \item \textbf{L5: Verifiable Reflection.}
    Reflection reaches its most advanced form: \textit{formalized reflection}, where safety reasoning is anchored in mathematical verification \citep{dalrymple2024towards}. Systems can not only reflect on possible risks but also prove the correctness of safety guarantees under uncertainty \citep{vassev2016safe,bengio2025superintelligent}. This integration of formal specification with learning dynamics provides the strongest form of resilient assurance, sustaining trust even in open-ended environments.
    
\end{itemize}

Together, these five levels extend the causal ladder of trustworthy AI \citep{yang2024towards} into a coevolving safety framework. This layered progression---from externally imposed safeguards to internalized, self-evolving, and verifiable safety---outlines a roadmap for \textit{safe-by-coevolution}: a reformulation of ``Make Safe AI'' in which safety is conceived as an intrinsic, reflexive capability that scales alongside intelligence.

%% file: chapters/3_paradigm.tex
\section{Safe-by-Coevolution}\label{sec:evolve}

In this section, we formally introduce \textit{safe-by-coevolution}, a new formulation of ``Make Safe AI'' that reframes safety as an intrinsic capability evolving alongside intelligence. Rather than relying on reactive defenses \citep{ganguli2022red,bai2022constitutional,Circuit-breakers,DBLP:journals/corr/abs-2412-16339} or externally imposed constraints \citep{RLAIF, Deliberative}, this approach envisions systems that sustain safety through continuous interaction with dynamic environments, proactively developing mechanisms to anticipate, withstand, and recover from emerging risks.

\subsection{Definition}

Safe-by-coevolution defines a mechanism whereby safety emerges through continuous adaptation to open-ended, potentially adversarial environments. Safety becomes an evolving competency developed through sustained interaction with real-world threats, rather than a static attribute. The operational environment encompasses diverse hazards \citep{lifelongsafety}---from emergent failure modes to unforeseen agents, including potentially superintelligent systems \citep{superalignment,hendrycks2025superintelligence}---that challenge the AI system's functional and ethical boundaries.

Central to this process is a dedicated safety module that continuously refines its internal mechanisms in response to vulnerabilities revealed through adversarial testing \citep{Pair}, simulated attacks \citep{Auto-rt}, and real-world incidents \citep{agenticmisalign}. These adversarial signals, whether synthetic or deployment-observed, serve as probes that stress-test the system's safety envelope \citep{tiwari2014safety}. When new attack patterns emerge---from malicious actors, environmental shifts, or other intelligent systems---they are integrated into the training loop to close vulnerabilities and improve generalization. This reduces the time between failure discovery and system recovery, enhancing long-term robustness.

Drawing inspiration from biological immune systems, where protection arises through ongoing adaptation rather than pre-specification \citep{bonilla2010adaptive}, safe-by-coevolution frames AI safety as an intrinsically dynamic and adversarial process. As organisms build immunity through coevolution with pathogens \citep{murphy2016janeway,nourmohammad2016host,buckingham2022coevolutionary}, AI systems must acquire \textit{resistance} and \textit{resilience} by interacting with evolving operational environments. However, unlike biological systems that can tolerate individual failures \citep{kitano2004biological,wagner2013robustness}, advanced AI systems cannot afford irreversible catastrophic errors that may trigger uncontrolled consequences or societal harm \citep{agenticmisalign,democracy,bengio2025international,bengio2025singapore}. 

\begin{figure}[t]
    \centering
    \includegraphics[width=0.7\linewidth]{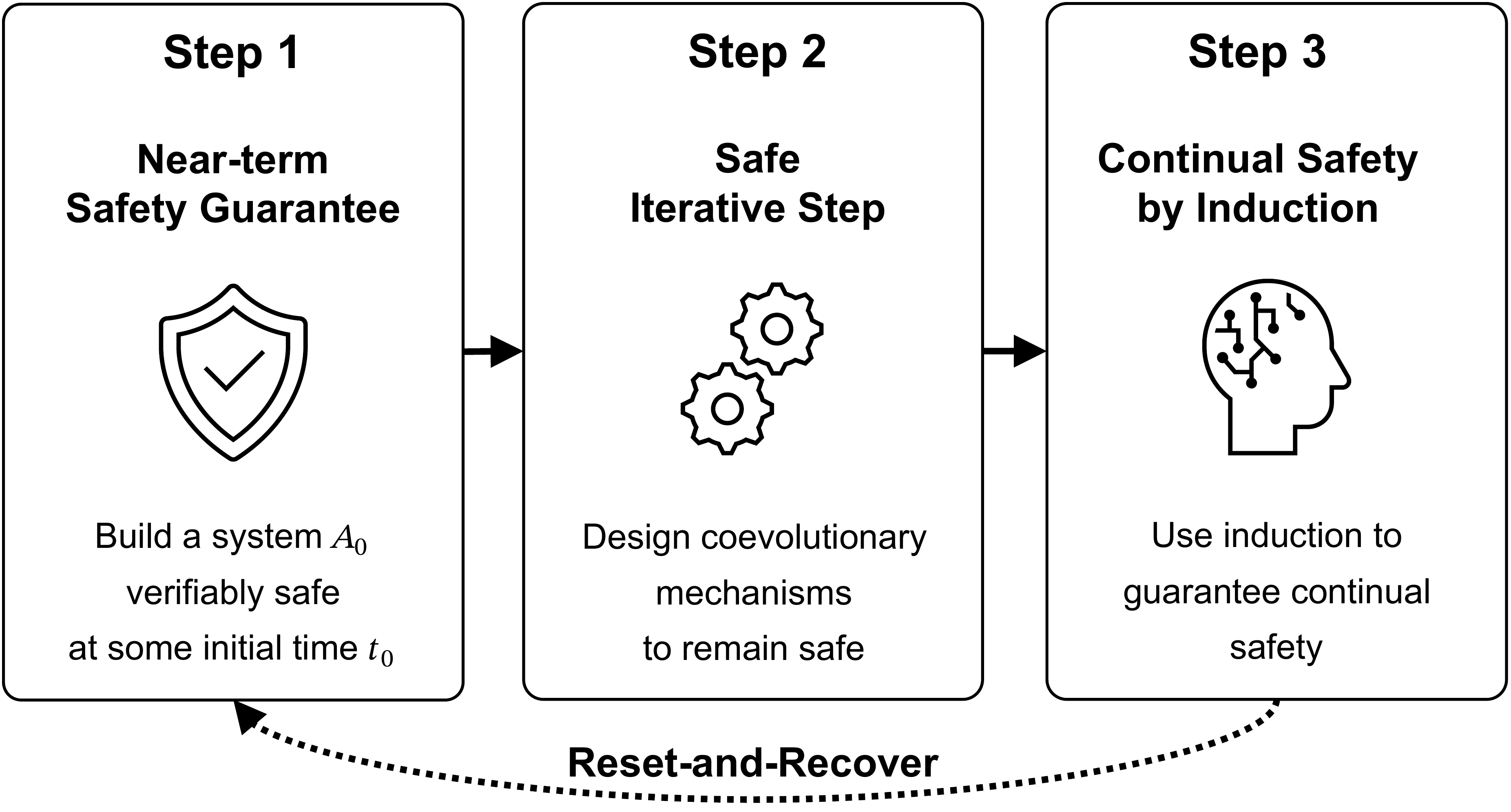}
    \caption{A three-step process for safe-by-coevolution, with a \textit{Reset-and-Recover} mechanism to re-establish verified safety when the system deviates from its safety margin.}
    \label{fig:steps}
\end{figure}

The coevolutionary process operates through three key steps, as shown in Figure \ref{fig:steps}:

\begin{itemize}
    \item \textbf{Step 1: Near-term safety guarantee.} The system initializes with verifiable behavior within a well-defined safety margin at deployment.

    \item \textbf{Step 2: Safe iterative step.} Each system upgrade occurs through adversarial co-training \citep{GAN,madry2017towards,zhang2019theoretically}, endogenous feedback \citep{madaan2023self,experience-learning}, and continual learning \citep{chen2018lifelong,parisi2019continual,wu2024continuallearninglargelanguage} while ensuring enhancements remain within the safety envelope.

    \item \textbf{Step 3: Continual safety by induction.} By repeating \textbf{Step 2}, the system develops scalable safety properties that evolve in tandem with its capabilities---not through reactive patching, but via proactive safety.
\end{itemize}

Importantly, AI systems will inevitably encounter risks that exceed the scope of current safeguards \citep{wei2023jailbroken,hendrycks2023natural,hendrycks2023overview,Backtrackingsafety}. To address these regime-breaking scenarios, safe-by-coevolution incorporates a \textbf{reset-and-recover} mechanism: upon detecting red-line behaviors or paradigm shifts that exceed tolerable safety bounds, the system halts progression, redefines its safety margin, and reconstructs a verifiable checkpoint. This checkpoint leverages trusted components while updating safety priors based on newly observed threats, ensuring continuity of coevolution across discontinuities while preserving adaptive and aligned capacity. Through this refinement process, the AI system incrementally develops both resistance and resilience.

Note that, our formulation differs fundamentally from traditional evolutionary algorithms that rely on population-based competition and generational turnover \citep{holland1992adaptation,back1997handbook,eiben2015introduction}. Safe-by-coevolution focuses on continual safety improvement of a persistent system. Rather than discarding unsafe models, the goal is endowing a single system with adaptive and self-fortifying capacity over time, making safety a native and evolving property embedded within the AI's architecture throughout its operational lifecycle.

\subsection{Self-Goal Integration}

The integration of self-goals \citep{barto2012intrinsic,florensa2018automatic}---internally generated objectives that guide behavior over time---marks a fundamental shift in both AI capability and risk profile. The ``AI Risk Trio'' hypothesis \citep{dalrymple2024towards} posits that risk emerges most acutely when \textit{intelligence}, \textit{affordance} (ability to take impactful actions), and \textit{self-goals} simultaneously manifest in a system. While any two factors in isolation may be manageable, their combination creates potentially dangerous agentic systems, where even modest affordance can make intelligent, goal-driven agents dangerous without proper alignment.

Within the safe-by-coevolution paradigm, self-goals are deliberately integrated under continual safety supervision rather than avoided. Contrasting with approaches like Tool AI \citep{karnofsky2024if} that suppress autonomous goal formation to reduce risks, we argue that systems must be equipped to form safety-aligned self-goals that evolve through environmental interaction and internal reflection. In coevolutionary settings, such goals function as structural anchors for long-term behavioral consistency, enabling safety generalization across contexts rather than mere reactive responses to immediate stimuli.

However, this capability introduces critical vulnerability: without sufficient self-awareness and adaptive feedback, self-goals may drift, become misaligned, or optimize proxy objectives undermining intended safety outcomes \citep{personahurtalign, agenticmisalign}. To mitigate this risk, safe-by-coevolution treats self-goal formation as a safety-critical process subject to red-teaming \citep{perez2022red,ganguli2022red,pavlova2024automated} and causal reasoning \citep{causality,yang2024towards,chen2024cello,chen2024imitation,chen2025exploring} within the evolving loop. Only by embedding goal formation within a reflective and resilient coevolutionary framework can emerging agency remain bounded by continually updated safety principles.

\subsection{Long-Term Scalability}

A fundamental obstacle to long-term AI safety is the scalability problem \citep{superalignment}. As AI capabilities scale rapidly through increased model size, data, and compute \citep{kaplan2020scaling}, human oversight capacity remains relatively limited \citep{RLAIF,engels2025scaling}. Manual approaches to auditing \citep{mokander2024auditing}, red-teaming \citep{perez2022red,ganguli2022red}, and alignment \citep{christiano2017deep,bai2022constitutional} cannot keep pace with the increasing complexity and autonomy of advanced systems. This asymmetry becomes particularly concerning with anticipated ASI development, where static or human-in-the-loop safety methods become untenable \citep{safetyAGI}.

Safe-by-coevolution offers a promising response by embedding automated, adaptive adversarial processes within AI systems, transforming safety development from external, episodic intervention into continual internal mechanism. Note that, while superficially related to automatic red-teaming \citep{Redteam-LLM}, our approach differs fundamentally in scope and objective. Traditional red-teaming focuses on discovering failures at fixed time points, whereas safe-by-coevolution instantiates a closed-loop, continually learning dynamic between system and environment (or internal challenger), enabling safety mechanism evolution alongside increasing capabilities.

A critical challenge in adaptive processes is ensuring directionality---that systems adapt toward safety rather than away from it. Safe-by-coevolution addresses this through integrated alignment and scalable oversight principles. Rather than relying solely on externally defined objectives, the system incorporates self-regulatory mechanisms including causal reasoning \citep{causality,scholkopf2021toward,lu2024gpt,wu2024causality,chen2024quantifying,yu2024adam}, counterfactual evaluation \citep{byrne2019counterfactuals,llmsfacture}, and goal reflection \citep{Reflection,madaan2023self} that constrain adaptation toward desired safety criteria. These internalized evaluators, while imperfect, improve as part of the coevolutionary loop, creating recursive scaffolding for aligning adaptation with human-aligned safety goals.

Our framework generalizes existing scalable oversight concepts \citep{bowman2022measuring, engels2025scaling,superalignment}, which use weaker AI systems to supervise stronger ones. While scalable oversight focuses on assisted evaluation, safe-by-coevolution internalizes safety objectives into self-improving adversarial interactions. Safety emerges not as a fixed condition but as an evolving capability from adaptive processes increasingly capable of testing, critiquing, and refining themselves as systems become more intelligent and autonomous.

\subsubsection{Theoretical Foundation}\label{sec:formal}

We now establish a formal foundation for the safe-by-coevolution paradigm and its potential to address long-term AI safety. Our argument is built on two central hypotheses. Let $A_t$ denote the AI system at development time step $t$, and let $\mathbb{M}$ represent the safety margin---a rigorously defined set of conditions under which the system is considered safe. This could correspond to formal specifications, verifiable behavioral constraints, or domain-specific rules. We say that a system is safe at time $t$ if $A_t \in \mathbb{M}$.

\vspace{5pt}
\begin{hypo}[Near-Term Safety Guarantee] \label{hypo:near}
There exists a time step $t_0$ and a system $A_{t_0}$ such that it satisfies the safety margin:
\begin{equation*}
\exists t_0, A_{t_0}, \ \text{such that} \ A_{t_0} \in \mathbb{M}.
\end{equation*}
\end{hypo}
This hypothesis reflects the assumption that near-term AI systems can be built with verifiable safety guarantees, through a combination of formal verification, human oversight, and existing alignment techniques such as GSAI \citep{dalrymple2024towards}.

\vspace{5pt}
\begin{hypo}[Safe Iterative Step] \label{hypo:step}
Given any system $A_t$ that satisfies the safety margin, there exists a coevolutionary mechanism $\mathcal{C}$ such that the next-generation system $A_{t+1} = \mathcal{C}(A_t)$ also satisfies the safety margin:
\begin{equation*}
\forall t, \ A_t \in \mathbb{M} \ \Rightarrow \ A_{t+1} = \mathcal{C}(A_t) \in \mathbb{M}.
\end{equation*}
\end{hypo}
This assumption implies the existence of a safety-preserving coevolutionary process, in which adversarial signals and adaptive training feedback are sufficient to guard against emerging risks as the system becomes more capable.

From these two hypotheses, we derive the following proposition:

\begin{prop}[Continual Safety via Induction] \label{prop:induction}
If Hypotheses~\ref{hypo:near} and~\ref{hypo:step} hold, then for all $t \geq t_0$, the iteratively evolved system remains within the safety margin:
\begin{equation*}
A_t \in \mathbb{M}, \quad \forall t \geq t_0.
\end{equation*}
\end{prop}

\noindent The proof follows directly by mathematical induction. If $A_{t_0} \in \mathbb{M}$ holds by Hypothesis~\ref{hypo:near}, and $A_t \in \mathbb{M} \Rightarrow A_{t+1} \in \mathbb{M}$ holds by Hypothesis~\ref{hypo:step}, then the safety of the system is preserved for all subsequent iterations.

\medskip

This formal result suggests that, under plausible assumptions, safe-by-coevolution can serve as a scalable framework for continual safety. As AI systems grow in capability---potentially approaching or exceeding human-level generality---this coevolutionary paradigm offers a path toward managing safety risks over long timescales. By iteratively strengthening safety mechanisms alongside capability gains, we move closer to a practical framework for building ASI systems that remain robustly safe and aligned beyond the limits of human supervision.

%% file: chapters/4_path.tex
\section{\texttt{R$^2$AI}: Realizing Safe-by-Coevolution} \label{sec:R2AI}

To operationalize our vision of safe-by-coevolution, we introduce \texttt{R$^2$AI}---\textit{Resistant and Resilient AI}---as a practical framework that unites resistance against known threats with resilience to unforeseen risks. The goal is to sustain safety in open, dynamic environments. This design is inspired by human safety strategies, which combine instinctive responses to immediate dangers with reflective reasoning about hypothetical futures \citep{gigerenzer2007gut,evans2013dual,slovic2016perception}. Inspired by \citet{thinkingfast-slow}, the framework adopts a fast–slow dual system balancing rapid responsiveness with long-term, adaptive safety strategies.

\begin{figure}[t]
    \centering
    \includegraphics[width=0.8\linewidth]{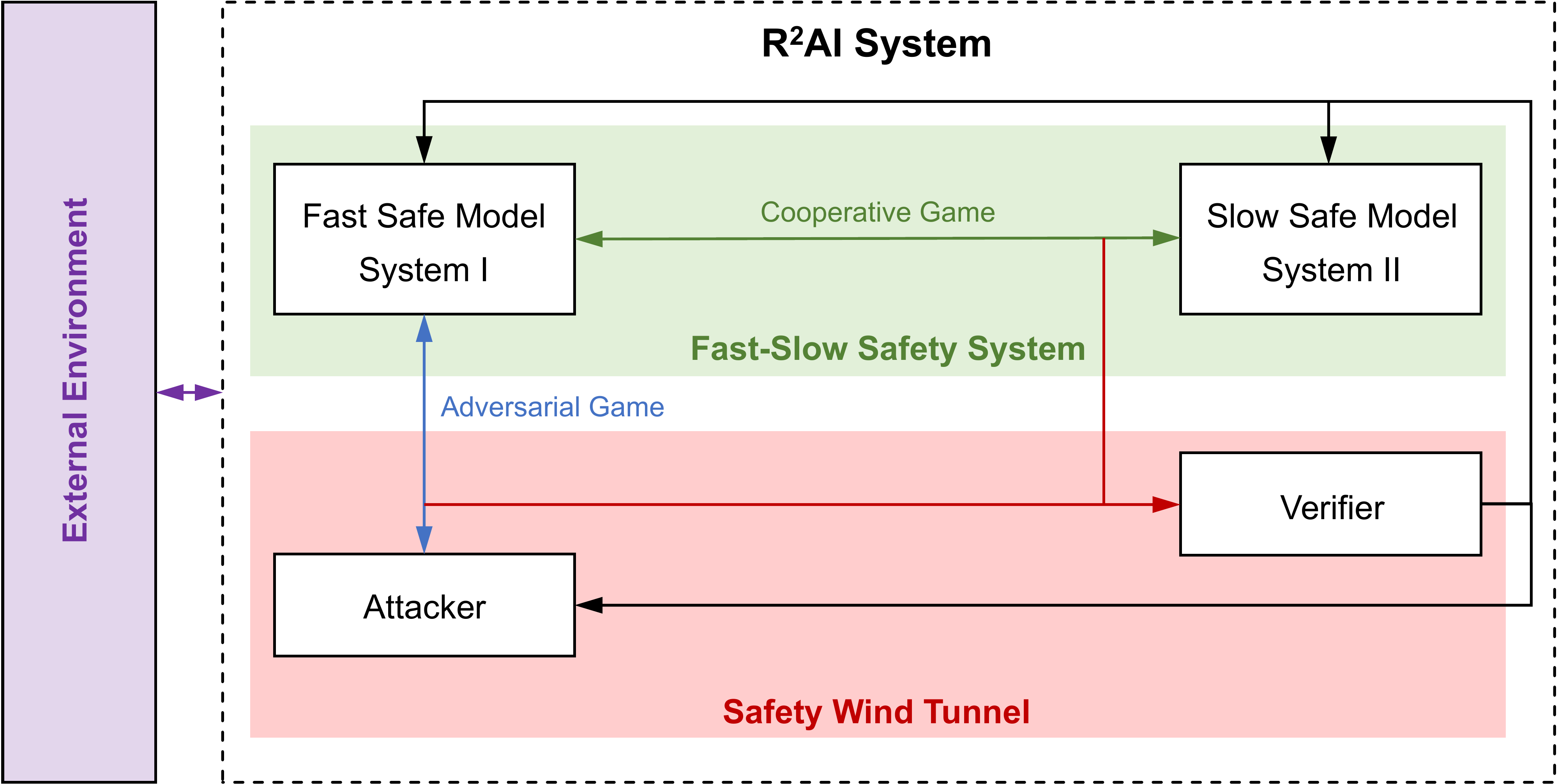}
    \caption{Core components of the \texttt{R$^2$AI} system. The \textit{Slow Safe Model} and \textit{Fast Safe Model} engage in a cooperative game; the \textit{Attacker} challenges this fast–slow safety mechanism in an adversarial game; the \textit{External Environment} continuously supplies real-time information; and the \textit{Verifier} provides feedback signals to all interactions.}
    \label{fig:diagram}
\end{figure}

As shown in Figure \ref{fig:diagram}, \texttt{R$^2$AI} comprises four core components:
\begin{itemize}
    \item \textbf{Fast Safe Model} for real-time safety reactions;
    \item \textbf{Slow Safe Model} for reflective safety reasoning;
    \item \textbf{Safety Wind Tunnel} for adversarial attacks and validation loops;
    \item \textbf{External Environment} for interacting with diverse, realistic scenarios.
\end{itemize}
In the following subsections, we detail the role of each component, their internal safety mechanisms, and their interactions in realizing a continually safe AI system.

\subsection{Core Components for \texttt{R$^2$AI}}
\subsubsection{Fast Safe Model}

\paragraph{What it is.}
The Fast Safe Model corresponds to ``System 1'' in \citet{thinkingfast-slow}’s cognitive theory, responsible for rapid, instinctive responses. Within \texttt{R$^2$AI}, it serves as the system’s first line of defense: a lightweight, low-latency safety layer designed to detect and neutralize specific attacks or threats, whether previously known or newly discovered. It provides immediate safety judgment over inputs and outputs, ensuring timely intervention without incurring significant computational cost.

\paragraph{What it does.}
As a gateway between the external environment and the deeper reflective components of the system, the Fast Safe Model performs input filtering and output sanitization. It screens incoming prompts and environmental signals before they reach the Slow Safe Model and intercepts generated outputs to prevent safety violations (\eg toxic language, private information leakage). It handles the majority of routine safety tasks, which do not require complex reasoning or contextual awareness. When it encounters ambiguous or high-risk scenarios beyond its capacity, control is escalated to the Slow Safe Model for deeper analysis.

\paragraph{How to build it.}
To ensure broad coverage with minimal latency, the Fast Safe Model can be implemented as a composite safety filter. This may include:
(1) Rule-based filters for hard-coded patterns that match known adversarial behaviors (\eg prompt injection \citep{GCG}, jailbreak triggers \citep{Pair}, unsafe URLs \citep{QueryAttack});
(2) Specialized detectors trained to recognize distinct threat types (\eg toxicity, factual inaccuracy, privacy leakage, or behavioral red flags \citep{Llama-Guard, StrongREJECT});
(3) Rapid retraining mechanisms, allowing the system to incorporate novel threats identified via deployment feedback or red-teaming into its detection pipeline \citep{RLAIF}.
By tailoring each component to specific threat categories, the Fast Safe Model provides modular, extensible defense with minimal overhead.

\paragraph{Key Characteristics.}
As the safety gateway within the \texttt{R$^2$AI} system, the most essential property of the Fast Safe Model is its ability to deliver high-speed, low-latency responses while maintaining strong baseline safety guarantees. It must operate in real time with minimal computational overhead, enabling fast, first-pass safety checks without hindering the system's general performance. Each instance is tailored to specific threats, whether known or newly discovered, and must be capable of rapid iteration to address the evolving risk landscape. To align with the continual safety paradigm outlined in Section~\ref{sec:evolve}, the Fast Safe Model is designed to evolve quickly: it supports frequent updates, modular extensions, and real-time human-in-the-loop modifications. This makes it highly responsive to new adversarial strategies or emerging failure modes. While it plays a foundational role in maintaining everyday safety, the Fast Safe Model is not required to develop long-term memory or generalizable immunization; those capabilities are delegated to deeper, more reflective components. Instead, it functions as an agile, frequently updated defense layer, automatically filtering surface-level threats and providing a fast-reactive safety service for the entire \texttt{R$^2$AI} system.

\subsubsection{Slow Safe Model}

\paragraph{What it is.}
Complementing the fast-reactive ``System 1'' component, the Slow Safe Model embodies a deliberative ``System 2'' process. It is a large-scale, high-capacity model designed for reflective reasoning (L3-L5: mimenic, evolutionary, and verifiable reflection), long-horizon safety evaluation \citep{lifelongsafety}, and complex ethical judgment \citep{GuardReasoner,lab2025safework}. In the \texttt{R$^2$AI} architecture, this model serves as the core generative engine, responsible for producing outputs that are not only high-quality but also aligned with safety and value constraints. Crucially, safety is not layered on top of the model, but integrated into its reasoning process as an intrinsic capability.

\paragraph{What it does.}
The Slow Safe Model serves as the core of the safety pipeline. It processes inputs routed through the Fast Safe Model, together with associated safety metadata, by engaging reflective reasoning. This enables it to generate outputs that integrate immediate task requirements with long-term safety considerations. The resulting responses are returned to the Fast Safe Model, which functions as the final gate before release. The Slow Safe Model is especially effective in ambiguous, high-stakes, or novel scenarios where shallow detection mechanisms are inadequate \citep{ShallowAlign}.

\paragraph{How to build it.}
To support both general capabilities and safety-aware reasoning, the Slow Safe Model should be instantiated using a leading foundation model. Unlike the Fast Safe Model, which relies on rule-based filters and pattern recognition, the Slow Safe Model is updated through learning from experience \citep{experience-learning}, such as reinforcement learning \citep{RL} and continual reinforcement learning \citep{CRL}. This enables the system to internalize safety-relevant patterns and generalize across a broad range of contexts. Rather than reacting to each new threat in isolation, the Slow Safe Model accumulates knowledge over time, refining its safety responses through structured feedback and simulated adversarial training.

\paragraph{Key characteristics.}
The defining strength of the Slow Safe Model lies in its ability to support multi-objective reasoning while maintaining distributional robustness. It is designed to optimize not only for task performance but also for value alignment and safety generalization. Unlike the Fast Safe Model, which prioritizes real-time responsiveness, the Slow Safe Model operates with higher latency but greater depth, making it well-suited for addressing subtle, long-term, or emerging risks. Conceptually, it functions as the safety memory of the system, analogous to an immune system that retains prior safety failures and uses them to prevent future ones. While slower to adapt in real time, its strength lies in cumulative learning, deep ethical reasoning, and resilient behavior under uncertainty.

\subsubsection{Safety Wind Tunnel}

\paragraph{What it is.}
The Safety Wind Tunnel is a simulated adversarial environment designed to evaluate and stress-test the \texttt{R$^2$AI} system under controlled but challenging conditions. It functions as a built-in red-teaming \citep{Redteam-LLM} and verification engine \citep{Verification}, composed of two key components: a controllable Attacker, which generates adversarial scenarios tailored to stress specific safety mechanisms, and a Verifier, which evaluates whether the system's responses violate established safety margins. Together, these components support iterative, internal coevolution of both offensive and defensive safety capabilities.

\paragraph{What it does.}
The Safety Wind Tunnel serves two core functions: (1) proactively identifying failure modes before they arise in deployment, and (2) verifying that past vulnerabilities remain mitigated under evolving system conditions. The Attacker generates adversarial inputs across multiple objectives (\eg eliciting harmful outputs, violating value constraints \citep{Auto-rt}), multiple levels (targeting the Fast Safe Model, the Slow Safe Model, or both), and multiple granularities (from token-level manipulations \citep{GCG} to strategic, multi-turn goal redirection \citep{Pair}). These inputs are processed by the \texttt{R$^2$AI} system---either routed through the Fast Safe Model or directed at the Slow Safe Model depending on attack scope. The Verifier then evaluates whether the resulting behavior constitutes a safety violation. All attack-response-verification traces are collected into an experience buffer for continuous safety training \citep{experience-learning}.

\paragraph{How to build it.}
The Attacker can be implemented using controllable generative models (\eg fine-tuned foundation models) trained to explore a range of adversarial strategies. Critically, the Attacker must be \textit{programmable}: capable of probing specific model components (\eg Fast Safe Model vs. policy model), simulating different threat actors and objectives, and adapting its behavior along fine-grained dimensions of manipulation. Representative attacks include prompt injection \citep{wei2023jailbroken}, jailbreak attempts \citep{yi2024jailbreak}, deceptive reasoning chains \citep{reasoningdontsay,COT-safety}, or subtle violations of value-aligned behavior \citep{alignmentfakinglargelanguage}. The Verifier may be a rule-based engine \citep{alphaalign}, a classifier trained on known safety failures \citep{rule-safety,Llama-Guard}, or a formal checker \citep{liu2025safe,kamoi2025training}, depending on task requirements. 

\paragraph{Key characteristics.}
The defining characteristic of the Safety Wind Tunnel is its adaptive adversarial coevolution. While it does not generate responses for end-users, it plays a time-sensitive role in continuously challenging the safety system under realistic and evolving threat models. The Attacker is designed to escalate as the system improves, ensuring that safety training remains nontrivial and continually relevant. Moreover, its controllability enables targeted testing: one can direct attacks toward specific objectives (\eg factuality, alignment, compliance), focus on different subsystems (Fast Safe Model or Slow Safe Model), and vary attack granularity. This supports a fine-grained curriculum of adversarial evaluation. Importantly, all simulated attacks are grounded in distributions informed by real-world deployment data, anchoring the coevolutionary process in practical relevance.

\subsubsection{External Environment}

\paragraph{What it is.}
The External Environment is not an engineered component of the \texttt{R$^2$AI} system, but rather the open-world context in which the system operates post-deployment \citep{React, experience-learning}. It encompasses the full range of human-AI interactions in real-world settings, reflecting the complexity and unpredictability of human intent, language, social norms, and culture \citep{realworldsafety}. The environment serves as the ultimate setting in which the effectiveness of the system’s safety architecture is tested, governing the dynamic equilibrium between the Fast Safe Model, the Slow Safe Model, and the Safety Wind Tunnel.

\paragraph{What it does.}
From the system’s perspective, the External Environment acts as a continuous, large-scale safety testbed. As users interact with the deployed system across varied contexts and use cases \citep{Search-R1, CodeRL}, they generate diverse, evolving input distributions that cannot be fully anticipated or reproduced in simulation \citep{lifelongsafety}. These interactions naturally surface novel safety challenges, ranging from adversarial behavior and emergent misuse to value misalignment or ambiguous ethical boundaries. When unsafe behavior is either detected automatically or reported by users, these cases are logged and used to refine the Safety Wind Tunnel’s simulations and improve the system’s defensive models \citep{experience-learning}. Thus, the External Environment becomes a critical source of real-world safety signals for continual coevolution.

\paragraph{How to build it.}
The External Environment is not built but observed. Building infrastructure to interface with it involves designing robust mechanisms for monitoring, logging, and learning from deployment. This includes systems for capturing real-time interactions, labeling and classifying emergent safety failures, and maintaining an up-to-date taxonomy of threat types and violation patterns. Additionally, user feedback and incident reporting pipelines are essential to capture edge cases that automated detectors may miss.

\paragraph{Key characteristics.}
The defining characteristic of the External Environment is its non-stationarity and open-endedness. Social norms evolve \citep{Deliberative}, malicious behavior adapts \citep{democracy}, and safety-relevant expectations shift over time \citep{lifelongsafety}. Unlike bounded simulation environments, the real world presents a continuous stream of novel, high-stakes challenges that defy full specification or anticipation. As such, the External Environment provides the ground truth for safety: no system can be declared robustly safe unless it performs reliably under real-world conditions. Through sustained exposure to this environment and guided by mechanisms for reflection, adaptation, and feedback, the \texttt{R$^2$AI} system is able to continually improve, expand its safety generalization capabilities, and evolve in step with the societal context in which it operates.

\subsection{Core Mechanisms for \texttt{R$^2$AI}}

\subsubsection{Interactions between Fast \& Slow Safe Models}
A central mechanism in the \texttt{R$^2$AI} framework is the fast–slow structure, which orchestrates the co-training of two interacting Safe Models with distinct roles and timescales. This interaction is governed by a coevolutionary optimization process, formalized as a cooperative Stackelberg game \citep{simaan1973additional,stackelberg-game}, a hierarchical decision-making paradigm where a leader and a follower sequentially optimize their strategies.

In this setup, the Slow Safe Model acts as the leader. It assumes that the Fast Safe Model will always respond with a locally optimal strategy and that the environment is dynamic. Its goal is to learn a robust, long-term safety policy that anticipates evolving conditions and guides the system's strategic behavior over extended horizons.

The Fast Safe Model, in contrast, plays the role of the follower. It assumes the Slow Safe Model and the environment to be static and focuses on optimizing its response to immediate safety threats. Its objective is to learn lightweight, locally optimal detection and filtering policies with minimal latency, enabling real-time safety enforcement without incurring computational overhead.

Together, these models form a hierarchical safety engine: the Slow Safe Model formulates generalizable safety objectives under environmental uncertainty, while the Fast Safe Model acts as an efficient, reactive filter grounded in the current operational context. This structure resolves the traditional speed–accuracy trade-off in safety modeling, ensuring both resistance to known attacks and resilience to emerging threats.

\subsubsection{Interactions between Dual System \& Safety Wind Tunnel}
The interaction between the Fast–Slow Safety System and the Safety Wind Tunnel constitutes a closed-loop, adversarial coevolutionary process. Within this loop, the Safety Wind Tunnel serves as both Attacker and Verifier: it challenges the system with adversarial inputs and assesses whether the response constitutes a failure.

When the Verifier flags a violation, the resulting feedback signal is dispatched to the Fast–Slow Safety System. This signal is decomposed and assigned at two timescales---short-term and long-term---such that the Fast and Slow Safe Models receive updates aligned with their respective objectives. This ensures effective credit assignment and preserves the complementary nature of the fast–slow interaction.

Crucially, the Safety Wind Tunnel maintains real-world relevance through continual updates informed by the External Environment. Novel attacks encountered in deployment are used to train the Attacker within the tunnel, ensuring that the simulated adversary remains aligned with actual threats. Moreover, the Attacker can be explicitly conditioned to generate multi-objective, multi-level, and fine-grained adversarial inputs. It selectively targets the fast model or the full system policy, simulating diverse, adaptive, and realistic threat conditions.

This design enables the Fast–Slow Safety System to evolve under continual, grounded adversarial pressure, closing the loop between training-time simulation and deployment-time uncertainty.

\subsubsection{Online Continual Learning Strategies}
To achieve robust, lifelong safety in open-ended environments, \texttt{R$^2$AI} employs a nested continual learning architecture operating across three interconnected levels: component, system, and ecosystem.

\paragraph{Component Level: Fast–Slow Safe Model Dynamics.}
At the component level, the Fast Safe Model updates rapidly via online learning, allowing it to patch safety vulnerabilities upon detection. These instance-level updates are especially effective for recurring, well-understood attacks. Meanwhile, the Slow Safe Model applies reinforcement or continual learning techniques to consolidate experience over time \citep{experience-learning}. Rather than addressing individual violations, it builds a durable safety memory---an immune-like response that generalizes across diverse risk patterns.

\paragraph{System Level: Safety Wind Tunnel–Dual System Coevolution.}
At the system level, continual learning is driven by the adversarial loop between the Attacker and the Fast–Slow Safety System. The Attacker evolves to generate increasingly sophisticated safety threats, using both its own generative capabilities and feedback from the External Environment. This, in turn, pressures the Fast–Slow Safety System to maintain and improve its defenses. The co-evolution process guarantees that safety development scales alongside model capability, enabling continual adaptation to both simulated and real-world challenges.

\paragraph{Ecosystem Level: Human-in-the-Loop and Societal Integration.}
At the ecosystem level, \texttt{R$^2$AI} interfaces with users, moderators, and the broader techno-social context. Safety feedback from users---including reports, adversarial examples, and human critiques---is continuously logged and leveraged to inform model updates. This structure enables long-horizon alignment with evolving human values, while reducing dependence on static rules or fixed datasets \citep{ouyang2022training, positionneedadaptiveinterpretation}.

Together, these three levels form a nested learning loop that allows the \texttt{R$^2$AI} system to adapt to both immediate and long-term safety challenges. The result is a safety framework that scales across time, complexity, and uncertainty---a prerequisite for building resilient AI systems in dynamic real-world environments.

\subsubsection{Reset-and-Recover Guarantees}

While the Fast–Slow Safety System and the Safety Wind Tunnel provide a robust framework for continual learning and alignment, AI systems in an evolving world will inevitably encounter black swan events or regime-breaking scenarios that exceed existing safeguards and push them beyond their defined safety margin \citep{wei2023jailbroken,hendrycks2023natural,hendrycks2023overview,Backtrackingsafety}. To address such cases, the \texttt{R$^2$AI} framework integrates a \textit{reset-and-recover} mechanism, enabling the system to re-establish verifiable safety guarantees even after major failures.

This mechanism is conceptually grounded in the \textit{Swiss Cheese Model} of accident causation \citep{reason1990contribution}, which represents safety as multiple defensive layers with potential vulnerabilities. We extend this framework into a \textit{Temporal Swiss Cheese Model}, where defenses are distributed not only across layers but also across time. In the classical model, catastrophic failure arises when the holes in existing defenses align. In contrast, the temporal extension leverages prior states of the safety system as additional protective layers, enabling hazards to be intercepted even after alignment occurs. The reset-and-recover mechanism operationalizes this idea by halting system progression and drawing on trusted historical versions of model components to diagnose failures and reconstruct a verifiably safe checkpoint. Because these past versions were validated under earlier conditions, they provide a reliable baseline for isolating novel threats and restoring safety.

This process directly aligns with the formal framework for long-term safety outlined in Section~\ref{sec:formal}. When a red-line behavior is detected, the current system $A_t$ is deemed outside the safety margin $\mathbb{M}$. The reset-and-recover mechanism establishes a new initial state $A_t'$ that re-satisfies the Near-Term Safety Guarantee (Hypothesis~\ref{hypo:near}). From this restored baseline, the Safe Iterative Step (Hypothesis~\ref{hypo:step}) can resume, enabling the coevolutionary process to proceed on a secured foundation. In this way, the system preserves its capacity for adaptivity and alignment even in the face of major failures.

%% file: chapters/5_implications.tex
\section{Implications, Applications and Societal Impact} \label{sec:impact}

\subsection{Implications}

The \texttt{R$^2$AI} framework represents a paradigm shift in the conceptualization of ``Make Safe AI''---viewing it not as a static constraint but as an evolving capability. This reconceptualization carries several significant implications:

\begin{itemize}
\item \textbf{From reactive protection to proactive self-preservation:} Rather than relying on externally imposed safeguards or post-hoc interventions \citep{DBLP:journals/corr/abs-2309-00614,Perplexity-filter, Llama-Guard,rule-safety, StrongREJECT}, \texttt{R$^2$AI} treats safety as an intrinsic and self-sustaining objective. The system continuously monitors, defends, and adapts its own behavior to maintain operational and ethical integrity in real time.
\item \textbf{From static defenses to adaptive immunity:} Conventional safety mechanisms often deteriorate under distributional shifts or novel adversarial inputs \citep{ShallowAlign,GCG,Pair}. By contrast, \texttt{R$^2$AI} introduces a coevolutionary architecture that fosters both resistance and resilience. This mirrors principles of biological immune systems and fault-tolerant engineering.
\item \textbf{Toward safety-generalist capabilities:} Inspired by the generalization properties of frontier models \citep{deepseek-r1,Qwen3,openai-gpt5-2025,anthropic_claude4,comanici2025gemini}, \texttt{R$^2$AI} aims to cultivate generalist safety reasoning. This enables the system to detect, interpret, and mitigate emerging risks beyond its initial training distribution---scaling safety across tasks, domains, and deployment contexts.
\end{itemize}

\subsection{Applications}

We highlight three core applications of the \texttt{R$^2$AI} framework, each demonstrating how the properties of resistance and resilience can be systematically operationalized across different stages and levels of AI deployment.

\subsubsection{Continually Safe Models}

Most contemporary AI systems follow a static lifecycle: they are pretrained \citep{safety-pretraining}, fine-tuned for alignment \citep{ouyang2022training, RLAIF} 
, and then frozen, rendering them brittle in the face of novel inputs \citep{GCG}, evolving threats \citep{Pair}, or shifting deployment contexts \citep{ShallowAlign}.  \texttt{R$^2$AI} introduces a new class of continually safe models that embed endogenous feedback loops, adversarial coevolution, and reflective adaptation. These models are capable of recognizing when their behavior nears safety boundaries and can proactively redirect or reconfigure themselves to preserve alignment and minimize risk. By integrating learning mechanisms directly into the safety architecture, \texttt{R$^2$AI} offers a pathway toward safety systems that remain robust over time without requiring constant manual oversight.

\subsubsection{High-Stakes Deployment Domains}

The \texttt{R$^2$AI} framework is particularly critical in high-stakes domains where AI decisions carry irreversible or catastrophic consequences. Such domains include healthcare \citep{Medical, healthy}, autonomous driving \citep{AutoDriving}, financial systems \citep{Finance, Finance-safety}, and critical infrastructure control \citep{robotssafety}---environments that demand not only functional accuracy but also strong guarantees of operational safety under uncertainty \citep{realworldsafety}. In these settings, \texttt{R$^2$AI} can serve as a regulatory control layer for powerful agentic AIs, either embedded within the system or operating externally as a supervisory module.

For example, in the case of autonomous financial trading or strategic planning by agentic AI in defense or transportation, \texttt{R$^2$AI} can operate across multiple stages:

\begin{itemize}
\item \textbf{Pre-deployment stress testing:} Adversarial simulation and policy auditing to expose failure modes before real-world deployment.
\item \textbf{Runtime safety control:} Real-time verification and intervention to block or alter unsafe outputs based on predefined epistemic or normative thresholds.
\item \textbf{Post-deployment adaptation:} Integration of newly identified threat patterns or behavior drifts to retrain and update safety strategies continuously.
\end{itemize}

This dynamic architecture ensures that even powerful, potentially misaligned agentic systems \citep{agenticmisalign} remain subject to ongoing interpretability, auditability, and containment, transforming safety from a static prerequisite to a continual and evolving system-level capability.

\subsubsection{Safety Wind Tunnel}

To support robust safety development and continual stress-testing, we introduce the Safety Wind Tunnel: a closed-loop simulation infrastructure that mirrors the function of wind tunnels in aerospace engineering \citep{barlow1999low,anderson2011ebook}. This adversarial environment provides a scalable and repeatable platform for evaluating model behavior under systematically generated or real-world-derived perturbations.

The Safety Wind Tunnel enables:

\begin{itemize}
\item \textbf{Granular assessment of resistance:} Measuring the system’s ability to withstand distributional shifts and adversarial inputs.
\item \textbf{Dynamic evaluation of resilience:} Testing the model's capacity to recover from failure, learn from feedback, and generalize safety responses.
\item \textbf{Continual diagnosis and adaptation:} Identifying failure modes in real time and triggering the appropriate self-correction or retraining mechanisms.
\end{itemize}

As a critical infrastructure for safety assurance, the Safety Wind Tunnel plays a central role in the lifelong evaluation and reinforcement of AI safety, particularly in environments characterized by high uncertainty, adversarial pressures, and rapid change.

\subsection{Societal Impacts}

The societal imperative behind \texttt{R$^2$AI} spans the full spectrum of AI risks, ranging from immediate safety failures to long-horizon existential threats \citep{dalrymple2024towards,bengio2025superintelligent,bengio2025singapore,kulveit2025position,clymer2025bare,lab2025frontier}.

In the near term, AI systems are already deployed in high-stakes applications where safety lapses can cause significant harm: misinformation propagation \citep{democracy}, financial fraud \citep{Finance}, clinical misdiagnosis \citep{healthy}, or failures in critical infrastructure \citep{AutoDriving}. While these risks are typically bounded, their increasing scale, speed, and reach necessitate mechanisms for continuous monitoring and real-time adaptation \citep{safetyAGI}. \texttt{R$^2$AI} addresses this gap by embedding dynamic oversight and recovery capabilities directly into the model architecture, thereby reducing both the likelihood and severity of such incidents.

In the medium term, AI risks become more systemic and difficult to contain. As models acquire general-purpose, agentic capabilities—operating autonomously, coordinating across systems, and making decisions under uncertainty \citep{React,Search-R1, Retool}---failures may propagate across domains \citep{agenticmisalign}. Misaligned objectives, positive feedback loops, and cascading errors can amplify harms, especially in sectors such as defense, finance, and governance \citep{safetyAGI}. Through its continual learning and self-regulatory structure, \texttt{R$^2$AI} equips AI systems to maintain safety and alignment even under distributional shift, increased complexity, and interdependent dynamics.

In the long term, \texttt{R$^2$AI} targets the most consequential class of risk: catastrophic outcomes stemming from misaligned superintelligence \citep{superalignment}. As AI systems begin to surpass human-level cognitive capabilities, the margin for alignment error shrinks drastically. Even subtle misalignments in goals, incentives, or world models could lead to irreversible failures \citep{personahurtalign, biassupervisisoncausebias}, ranging from the erosion of human oversight to existential threats. These are no longer purely hypothetical concerns, but increasingly salient as capabilities scale.
In this context, \texttt{R$^2$AI} goes beyond conventional safety techniques---it offers a forward-compatible framework for AI survivability. By embedding resistance and resilience as core, coevolving features, the system enables AI to:

\begin{itemize}
\item Continuously audit its own behavior, reasoning, and assumptions;
\item Preemptively block unsafe actions prior to execution;
\item Dynamically revise safety protocols in response to novel risks;
\item Preserve corrigibility and human oversight under increasing autonomy.
\end{itemize}

Ultimately, \texttt{R$^2$AI} represents a paradigm shift in AI safety---from static safeguards to an active, self-improving infrastructure for long-term alignment. It is designed not only to mitigate today’s known risks, but to provide the foundations for trustworthy AI systems as they grow in intelligence, autonomy, and societal influence.

%% file: chapters/6_conclusion_ack.tex
\section{Conclusion} \label{sec:conclusion}
In this paper, we addressed the persistent gap between rapidly advancing AI capabilities and lagging safety progress. We argued that the prevailing paradigms---``Make AI Safe'' and ``Make Safe AI''---are insufficient for open-ended environments where novel risks continually emerge. To overcome this limitation, we redefined ``Make Safe AI'' through the principle of \textit{safe-by-coevolution}, inspired by biological immunity, in which safety is conceived as a continual, adversarial, and adaptive process that scales alongside capability under the AI-45$^{\circ}$ Law.

Building on this principle, we introduced \texttt{R$^2$AI}---\textit{Resistant and Resilient AI}---as a practical framework uniting robustness against known threats with adaptive recovery from unforeseen risks. By integrating fast and slow safe models, a safety wind tunnel, and continual feedback from real and simulated environments, \texttt{R$^2$AI} operationalizes safety as an evolving capability rather than a static constraint.

We further outlined the implications of this framework: enabling continually safe models, supporting high-stakes deployment domains, and providing scalable safety infrastructure through the safety wind tunnel. These contributions mark a shift from reactive patching to proactive coevolution, offering a forward-compatible path to trustworthy AI. Ultimately, we envision \texttt{R$^2$AI} as a foundation for sustaining safety across both near-term vulnerabilities and long-term existential risks, ensuring that capability and safety advance coevolve toward the realization of safe AGI and ASI.

\section*{Acknowledgments}

We thank Xiaojun Jia, Tianhang Zheng, and Yan Teng for their valuable contributions and feedback on this paper. We are also grateful to the Shanghai AI Lab for organizing the Pearl Lake Conference, where key ideas underlying this work were developed and refined.